\documentclass[nohyperref]{article}

\usepackage{microtype}
\usepackage{graphicx,booktabs,array}
\usepackage{makecell}
\usepackage{subfigure}
\usepackage{booktabs} 

\usepackage{hyperref}

\usepackage[accepted]{icml2022}

\usepackage{amsmath}
\usepackage{amssymb}
\usepackage{mathtools}
\usepackage{amsthm}
\usepackage{bbm}

\usepackage[capitalize,noabbrev]{cleveref}

\theoremstyle{plain}
\newtheorem{theorem}{Theorem}[section]

\newtheorem{lemma}[theorem]{Lemma}

\theoremstyle{definition}

\theoremstyle{remark}

\usepackage[textsize=tiny]{todonotes}

\icmltitlerunning{Physics-informed disentanglement of multimodal data}

\begin{document}

\twocolumn[
\icmltitle{Unsupervised physics-informed disentanglement of multimodal data\\ for high-throughput scientific discovery}



\icmlsetsymbol{equal}{*}

\begin{icmlauthorlist}
\icmlauthor{Nathaniel Trask}{xxx}
\icmlauthor{Carianne Martinez}{yyy,comp}
\icmlauthor{Kookjin Lee}{comp}
\icmlauthor{Brad Boyce}{zzz}

\end{icmlauthorlist}

\icmlaffiliation{xxx}{Center for Computing Research, Sandia National Laboratories, Albuquerque, NM, USA}
\icmlaffiliation{yyy}{Applied Information Sciences Center, Sandia National Laboratories, Albuquerque, NM, USA}
\icmlaffiliation{zzz}{Materials, Physical, and Chemical Sciences Center, Sandia National Laboratories, Albuquerque, NM, USA}
\icmlaffiliation{comp}{School of Computing and Augmented Intelligence, Arizona State University, USA}

\icmlcorrespondingauthor{Nathaniel Trask}{natrask@sandia.gov}

\icmlkeywords{Multimodal machine learning, physics-informed machine learning, variational inference, variational autoencoders, fingerprinting, mixture of experts}

\vskip 0.3in
]



\printAffiliationsAndNotice{}  

\begin{abstract}
We introduce physics-informed multimodal autoencoders (PIMA) - a variational inference framework for discovering shared information in multimodal scientific datasets representative of high-throughput testing. Individual modalities are embedded into a shared latent space and fused through a product of experts formulation, enabling a Gaussian mixture prior to identify shared features. Sampling from clusters allows cross-modal generative modeling, with a mixture of expert decoder imposing inductive biases encoding prior scientific knowledge and imparting structured disentanglement of the latent space. This approach enables discovery of fingerprints which may be detected in high-dimensional heterogeneous datasets, avoiding traditional bottlenecks related to high-fidelity measurement and characterization. Motivated by accelerated co-design and optimization of materials manufacturing processes, a dataset of lattice metamaterials from metal additive manufacturing demonstrates accurate cross modal inference between images of mesoscale topology and mechanical stress-strain response.
\end{abstract}

\section{Motivation}

Many scientific and engineering datasets are multimodal, necessitating the fusion of disparate sources and datatypes for informed analysis.  For example, in the realm of process optimization for materials manufacturing, processes ranging from microelectronic fabrication to metal additive manufacturing involve a myriad of process settings along with in-process and post-process measurements \cite{liu2012micromachined,sochol20183d}. Moreover, automated high-throughput characterization methods are increasingly generating large, rich, multimodal datasets, fueled by advances in robotics and automation \cite{boyce2019autonomous}. 
Many scientific datasets admit \textit{fingerprints}: easily measurable signals which correlate with a difficult to measure underlying physical process. The hunt for exploitable fingerprints extends beyond material science \cite{isayev2015materials}, and can be found in all science and engineering domains ranging from quantum mechanics \cite {chakraborty2021qm} to climate change \cite{hasselmann1997multi,hegerl2007understanding}. Rapid datasets designed to detect fingerprints may potentially serve as a surrogate for, or in conjunction with, bespoke experiments capturing high-fidelity modalities. 
Accordingly, we aim to discover comprehensive fingerprints constructed from the weighted integration of several disparate data sources, each with unique fidelity, sparsity, and spatiotemporal resolution.  The \textit{multimodal scientific data} under consideration includes both physical and simulated data and differs from the text/audio/video modalities commonly considered in the multimodal literature \cite{baltruvsaitis2018multimodal}, providing the opportunity to impose physics-based inductive biases and move beyond purely data-driven linear techniques such as principle component analysis typically used for fingerprint detection. 

Herein, we present a novel variational inference framework for synthesizing multimodal scientific data with the aim of \textit{cross-modal inference}; if one can reliably perform generative modeling of a high-fidelity but slow measurement from a low-fidelity but fast fingerprint, high-throughput experimentation and material characterization are no longer infeasible. Such applications however mandate an unsupervised approach, as costly human-in-the-loop data labelling precludes high-throughput testing.

\begin{figure*}[h]
  \includegraphics[width=0.99\textwidth]{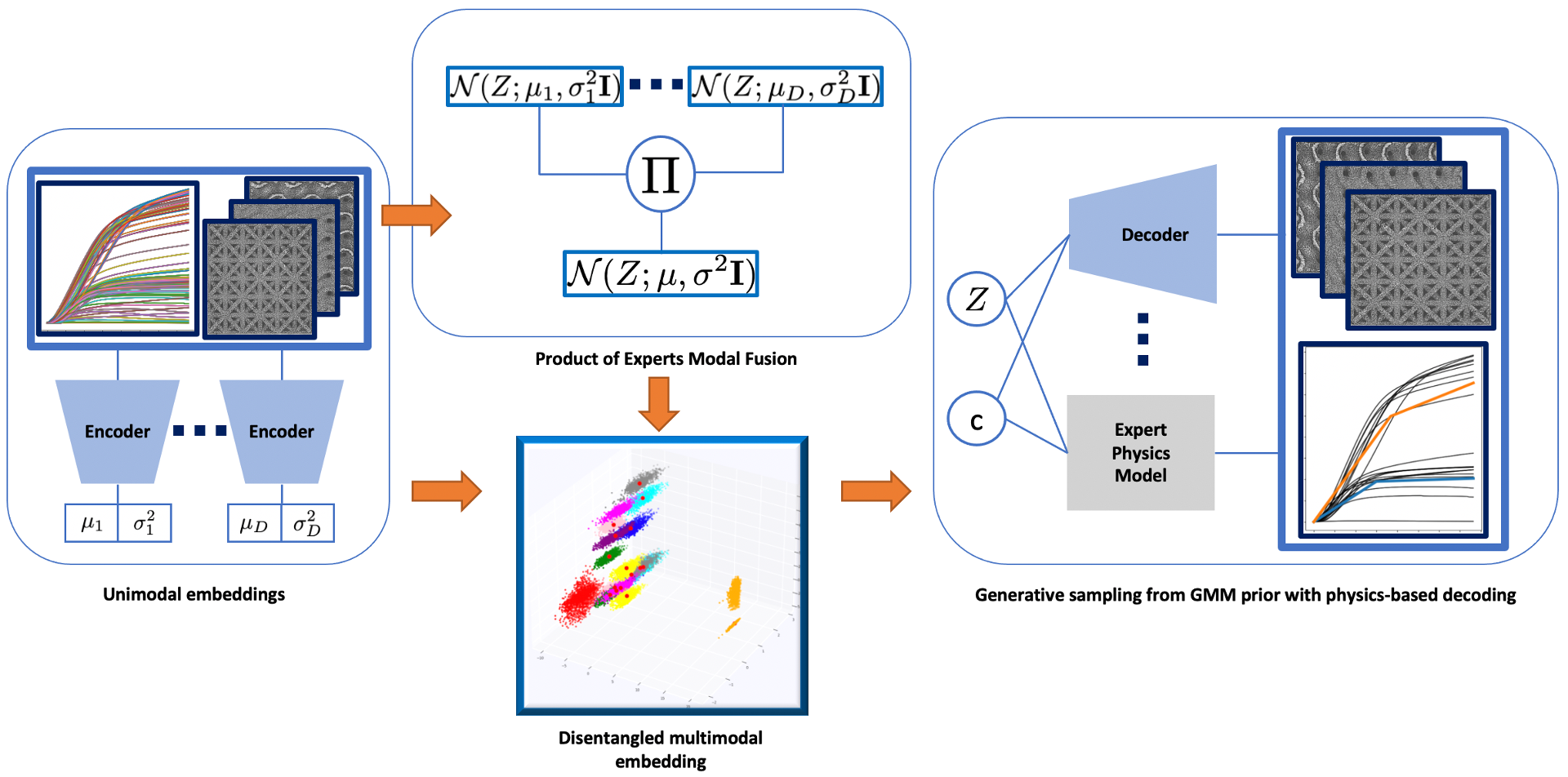}
  \caption{Individual modalities are encoded into Gaussian distributions in a shared latent space. During training the posterior is sampled from a product of experts distribution fusing complementary information into a shared multimodal Gaussian distribution. A Gaussian mixture prior parameterizes clusters encoding cross-modal shared information. Sampling from mixture components provides generative models using either black-box decoders or expert physics models encoding prior physics knowledge. To facilitate cross-modal generative inference, unimodal embeddings are trained to reproduce the multimodal embedding, allowing inference of $p(c|X_i)$. Shown here, an expert strain-hardening plasticity model allows two types of cross-modal inference: costly measurements of stress-strain response may be inferred from high-throughput imaging of lattice topology, or generative microstructural images can be provided to suggest microstructure which correlate with a given stress/strain measurement.}
  \label{roadmap}
\end{figure*}

Concretely, cross-modal inference corresponds to training jointly across modalities $X_1,...,X_D$ in a manner that supports generative sampling of individual modalities $p(X_i|X_j)$ for $i \neq j$. We achieve this in a variational inference setting by combining the following algorithmic contributions (Figure \ref{roadmap}): \textbf{1.} encoding data into unimodal embeddings $q(Z|X_i)$ and applying a product of experts (PoE) model to fuse data into a multimodal posterior $q(Z|\mathbf{X}) = \Pi_i q(Z|X_i)$; \textbf{2.} adopting a Gaussian mixture prior $p(Z|c)$ to determine latent clusters shared across modalities; and \textbf{3.} decoding with a physics-informed mixture of experts (MoE) model $p(X_i|c,Z)$ to impose inductive biases. For scientific settings, the expert model provides a critical new means of fusing experimental audiovisual data with traditional scientific models; rather than considering generalized linear models commonly used in MoE \cite{jordan1994hierarchical}, we may incorporate parameterized physical models, surrogates or simulators for the system under consideration. These ingredients are designed to yield an ELBO loss with closed form expressions for requisite integrals and is amenable to a novel expectation maximization strategy to fit clusters and experts. In concert, this architecture produces fingerprints in the form of latent clusters spanning modalities, with cross-modal estimators allowing inference of cluster membership for a single modality.

For unsupervised learning, several works apply variational autoencoders (VAE) to seek latent \textit{disentangled representations} of data which admit efficient separation into meaningful classes \cite{burgess2018understanding,chen2018isolating,locatello2019challenging,kim2018disentangling}. While desirable from an interpretability and accuracy perspective, such representations are often challenging to reliably discover in the absence of labels. The complementary information available in multimodal data has been shown to provide multiple pathways to disentanglement; a human may be unable to distinguish an image of a one and a seven, but if the digit is read aloud there is no confusion. However, the fact that scientific data is governed by physical models potentially allows the expert model to extract more information than purely data-driven approaches - known physics encodes the generative process and therefore imposing even a low-fidelity physical model as inductive bias may provide substantial disentanglement. Our tests demonstrate that the combination of PoE, GMM, and expert models provide not only disentangled clusters, but also an ordering of clusters in latent space reflecting information shared across modalities. For examples with unambiguous class ownership, we demonstrate 100\% test accuracy when classifying data into clusters. We anticipate these physics-based disentangled representations will enable future causal analysis of large high-throughput datasets.

\subsection{Relationship to prior literature}
This work draws from several thematic bodies of literature. The non-exhaustive list below denotes those works which have most informed our approach as well as provide overviews for the recent state-of-the-art.
\paragraph{Gaussian mixture embeddings} For deep unsupervised clustering, several works replace the standard normal prior from \cite{Kingma2014auto,rezende2014stochastic} with a Gaussian mixture model (GMM) to facilitate disentanglement and provide an explicit parameterization of clusters \cite{dilokthanakul2016deep, jiang2016variational,rao2019continual,lee2020meta}. Each modality in the multi-modal prior distribution is expected to provide disentangled latent representations of data which admit an explicit parameterization of class distributions. The current work is most similar to VaDE \cite{jiang2016variational} in its use of mean-field distributions to obtain a separable ELBO, and Bayesian estimator for $q(c|X)$. This work builds upon VaDE by incorporating multimodal data inputs while maintaining computational tractability of the ELBO, as well as employing clusters to decode into physics-informed MoE models.

\paragraph{Disentanglement} Another line of research is to extract latent disentangled representations into different factors of variations in data using VAEs. Earlier works such as $\beta$-VAE \cite{higgins2017betavae} and Annealed VAE \cite{burgess2018understanding} introduce additional weighting parameters to the KL divergence term of the original VAE ELBO loss. In Factor VAE \cite{kim2018disentangling} and $\beta$-TCVAE \cite{chen2018isolating} the ELBO is further decomposed to derive and penalize the total correlation to promote disentanglement in learned representations. For our purposes however, without an explicit parameterization of the cluster distributions to condition off of, it is not possible to introduce a physics-informed expert MoE model.

\paragraph{Multimodal inference} Generative modeling from multimodal data can be broadly categorized into either conditional generative models \cite{sohn2015learning,pu2016variational} which directly learn conditional cross-modal distributions $p(X_i|X_j)$, or joint models  \cite{suzuki2017joint,vedantam2018generative,wu2018multimodal}, which explicitly learn joint distributions that learn $p(Z,X_1,...,X_D)$. We pursue the later as \cite{wu2018multimodal} has been shown to provide better description of the underlying data distribution. We pursue the strategy used by works such as joint multimodal VAE \cite{suzuki2017joint} and joint VAE \cite{vedantam2018generative}, where a a joint inference network $q(Z|X_1,X_2)$ is trained, followed by training of two additional unimodal inference networks $q(Z|X_1)$ and $q(Z|X_2)$ which handle missing data at test time. The unimodal inference networks are trained to either match the joint inference network or to maximize an ELBO derived to perform unimodal variational inference. More recently, MVAE \cite{wu2018multimodal} and MMVAE  \cite{shi2019variational} were proposed to model the joint posterior as a product of experts (PoEs) and a mixture of experts (MoEs). Most recently, MoPoE-VAE \cite{sutter2021generalized} proposed a new ELBO formulation, which generalizes ELBO formulations derived from PoEs and MoEs. Our encoder bears similarities to MMVAE, MoPoE-VAE, and PoE, while preserving a computationally tractable closed form ELBO when combined with the GMM prior.

\paragraph{Physics-informed ML and fingerprinting} 
Substantial works in recent years have focused on introducing physics into either solving partial differential equations (PDEs) or for building surrogates, typically introducing a PDE residual regularizer in \textit{physics-informed neural networks} \cite{lagaris1998artificial,raissi2019physics} or by embedding physics directly into network architecture in \textit{structure-preserving ML} \cite{trask2020enforcing}. Such tools can be combined to provide parametric surrogates of simulations which can perform real-time inference over a database of parameterized PDE solutions \cite{lu2019deeponet,wang2021learning,mao2021deepm}. This paper provides a framework to fuse either these physics-informed surrogates or simpler empirical models together with experimental data. In contrast to traditional tools for fingerprinting which rely on purely data-driven techniques like PCA \cite{hasselmann1997multi,hegerl2007understanding}, the current framework provides a means to incorporate domain expertise into fingerprints tailored toward a scientific task.

\paragraph{Major contributions:}
\begin{itemize}
    \item Novel fusion of PoE w/ Gaussian mixture to obtain parameterized cluster ``fingerprints" for downstream data analysis and high-throughput diagnostic tasks.
    \item Multimodal embedding allows cross-modal inference while preserving closed form expressions for expectations in ELBO.
    \item Mixture of experts decoding allows incorporation of interpretable inductive biases by assuming model form describing scientific processes. Potential for embedding physics-informed surrogates or simulators.
    \item Improvements over SotA unimodal unsupervised techniques.
    \item Disentanglement of clusters into structured latent space exposing relationships across modalities.
\end{itemize}

\section{Framework construction}

Given individual modalities $X_i \subset \mathbb{R}^{d_i}$, we partition the set of all modalities $\mathcal{M}=\left\{X_1,...,X_D\right\}$ into $\mathcal{M}_{DD}$ consisting of images/videos/audio amenable to purely data-driven modeling, and $\mathcal{M}_S$ consisting of scientific modalities amenable to expert modeling. For each $X_i \in \mathcal{M}$ we seek an embedding $Z\in\mathbb{R}^l$ in latent dimension $l<<{d_i}$. Assuming a categorical variable $c$ clustering data into $C$ clusters in latent space, our variational autoencoder amounts to introducing parameterized prior $p$ and posterior $q$ distributions that maximize the following ELBO loss:
\begin{equation}
    \mathcal{L} = \mathbb{E}_{q(Z,c|X_1,...,X_D)}\left[\log \frac{p(X_1,...,X_D,Z)}{q(Z,c|X_1,...,X_D)} \right].
\end{equation}
We further assume separability of both prior and posterior:
\begin{align}
    p(X_1,...,X_D,Z,c) = \left(\Pi_{i=1}^D p(X_i|Z,c)\right) p(Z|c) p(c), \\
    q(Z,c|X_1,...,X_D) = q(Z|X_1,...,X_D)q(c|X_1,...,X_D). 
\end{align}
Our framework consists of four components: \textbf{1.} unimodal deep encodings with a product of experts (PoE) multimodal fusion, \textbf{2.} a mixture of Gaussians prior,  \textbf{3.} a mixture of experts decoding of modalities $X\in\mathcal{M}_S$, and \textbf{4.} unimodal encoders for cross-modal inference. We introduce each component sequentially, derive a closed form expression for the ELBO, and introduce an expectation maximization assignment of clusters and expert models.

\subsection{Multi-modal embedding}
Assuming the unimodal embeddings may be modeled as multivariate Gaussians with diagonal covariance, we obtain posterior probabilities $q(Z_i|X_i) = \mathcal{N}(Z_i;\mu_i,\sigma^2_i \mathbf{I})$, with mean and covariance provided by the set of neural networks
\begin{equation}
    [\mu_i,\sigma^2_i] = F_i(X_i; \theta_i),
\end{equation}
where $\theta_i$ denotes trainable weights and biases. For this work we consider a simple class of 1D/2D convolutional encoders, whose architecture is provided in Appendix \ref{app_archhyp}.

To estimate $q(Z|X_1,...,X_D)$ in the ELBO, it follows from Bayes' rule and pairwise independence that
\begin{equation}
    q(Z|X_1,...,X_D) = q(Z)^{1-D} \Pi_{i=1}^D q(Z|X_i),
\end{equation}
so that the posterior is a scaled product of individual modalities. To obtain closed form expressions for the ELBO later, we assume
\begin{equation}
    q(Z|X_1,...,X_D) \propto \Pi_{i=1}^D q(Z|X_i).
\end{equation}
The product of Gaussian distributions is again Gaussian, yielding the multimodal distribution:
\begin{align}\label{productPrior}
    q(Z|X_1,...,X_D) = \mathcal{N}(\mu, \sigma^2 \mathbf{I}),\\
    \sigma^{-2} =  \sum_{i=1}^D \sigma_i^{-2}, \qquad \frac{\mu}{\sigma^2} = \sum_{i=1}^D \frac{\mu_i}{\sigma_i^2},
\end{align}
which may be sampled during training using the reparameterization trick: by sampling $\epsilon \sim \mathcal{N}(0,\mathbf{I})$ and calculating $Z = \mu + \epsilon \odot \sigma$, we may back-propagate through the random node $Z$ into the unimodal encoders, where $\odot$ denotes the Hadamard product. 

\subsection{Gaussian mixture prior and expert decoding}
We adopt a simple Gaussian mixture prior, modeling
\begin{align}
    p(c) = Cat(c|\mathbf{\pi}), \label{eq:p_c}\\
    p(Z|c) = \mathcal{N}(\mu_c \label{eq:p_z_given_c}, \sigma^2_c\mathbf{I}),\\
    p(X_i|Z,c) = \mathcal{N}(\hat{\mu}_i, \hat{\sigma}^2_i\mathbf{I}) \label{eq:p_x_given_z_c}.
\end{align}
To ensure a positive $\mathbf{\pi}$ that sums to unity, we parameterize it as the softmax of a trainable vector. To decode $X_i \in \mathcal{M}_{DD}$, we employ a neural network with parameters $\hat{\theta_i}$
\begin{equation}
    [\hat{\mu}_i, \hat{\sigma}^2_i] = G_i(X_i; \hat{\theta_i}).
\end{equation}
For $X_i \in \mathcal{M}_S$, we assume an expert model $p(X_i|c) = \mathcal{N}(\mathcal{E}(t;\hat{\theta}_c),\hat{\sigma}^2_{c} \mathbf{I})$, where $t$ is an independent variable and $\hat{\theta}_c$ denotes expert parameters associated with each cluster. The specific choice of $\mathcal{E}$ will be problem dependent and specified in the experiment section.

Because $p(X_i|Z,c)$ admits interpretation as a mixture of experts model \cite{jordan1994hierarchical},  we obtain closed form expressions for the mean and variance to facilitate postprocessing and uncertainty quantification:
\begin{align}
    \mathbf{E}[X_i] = \sum_c \pi_c \mathcal{E}(t;\hat{\theta}_c),\\
    \text{Var}[X_i] = \left(\sum_c \pi_c (\hat{\sigma}_c^2 - \mathcal{E}(t;\hat{\theta}_c)^2)\right) - \mathbf{E}[X_i]^2.
\end{align}

In practice, $\mathcal{E}$ may take a variety of forms and its judicious selection imparts significant prior knowledge. We consider in this work simple generalized linear models and, for the mechanical data, an empirical linear strain-hardening model. In general, these could range from empirical engineering correlations obtained from e.g. dimensionless analysis or singular perturbation theory, to analytic parametric solutions to PDE based models, or to parametric physics-informed ML surrogates/reduce order models (see e.g. \cite{lu2019deeponet,wang2021learning,mao2021deepm,trask2020enforcing}).

\subsection{ELBO loss and EM minimizer}
A modification of the derivation in \cite{jiang2016variational} to account for multimodality yields the closed form for the single sample ELBO:
\begin{align*}
    \mathcal{L}_d = - \sum_{X_i \in \mathcal{M}_{DD}} || X_{i} - \hat{\mu}_{i} ||^2\\
    - \sum_{X_i \in \mathcal{M}_{S}}\sum_{t_n}\sum_c \gamma_c \left(\log \hat{\sigma}^2_c + \frac{(X_{i,n} - \mathcal{E}(t_n;\hat{\theta}_c))^2}{\hat{\sigma_c}^2}\right)\\
    -  \sum_c \sum_{j} \gamma_c  \left( \log \sigma_{c,j}^2 + \frac{\sigma^2_j}{\sigma^2_{c,j}} + \frac{ (\mu_j - \mu_{c,j})^2 }{\sigma^2_{c,j}} \right) \\
    +2 \sum_c \gamma_c \log \frac{\pi_c}{\gamma_c}+ \sum_j \left(1 + \log \sigma^2_j\right),
\end{align*}
where $||\cdot||$ denotes the $\ell_2-$norm, subscripts denote scalar components of tensors, $\gamma_c$ is the posterior distribution
\begin{equation}\label{eq:gamma_posterior}
    \gamma_c = p(c|Z) = \frac{\pi_c p(Z|c)}{\sum_{c'}\pi_{c'} p(Z|c')},
\end{equation}
we estimate $q(c|X_1,...,X_d)=p(c|Z)=\gamma_c$ following \cite{jiang2016variational}, and we have taken $\hat{\sigma_c}=1$.  The derivation may be found in Appendix \ref{app_elbo}. We seek the minimizer of this loss over the entire data set: $\mathcal{L} = -\sum_d \mathcal{L}_d$.

In standard expectation-maximization fashion, we note that for fixed value of $\gamma_c$, the variation of $\mathcal{L}$ with respect to the cluster centers $\mu_c$ yields the global minimizer
\begin{equation}\label{EMmean}
    \mu_c = \frac{\sum_d \gamma_{cd} \mu_d}{\sum_{d} \gamma_{cd}},
\end{equation}
where $\mu_d$ and $\gamma_{cd}$ denote the encoded mean and posterior $p(c|X_1,...,X_D)$ of the $d^{th}$ data point, respectively.

To facilitate batched access to large datasets, this may be calculated incrementally following the streaming algorithm outlined in Algorithm \ref{alg:training}, alternating between an EM update for $\mu_c$ followed by an Adam update \cite{kingma2014adam} of the remaining variables. 

A weighted least squares problem for the optimal expert model parameters may be similarly obtained by taking the variation of the ELBO with respect to $\hat{\theta}_c$:
\begin{equation}
    \hat{\theta}_c = \underset{\theta'}{\text{argmin}} \sum_d \sum_{t_n} \gamma_{cd} \left(X_{i,n} - \mathcal{E}(t_n;\hat{\theta}')\right)^2.
\end{equation}
Efficient solution of this nonlinear least squares problem at each epoch will be dependent upon the problem-specific expert model and data stream, and to perform batching may require a streaming technique such as recursive least squares or Kalman filtering \cite{cioffi1984fast}. For simplicity, we update $\hat{\theta}_c$ with Adam in this work but note that solving this at each epoch to ensure the expert model provides a best fit to the current partitions is likely to provide substantial improvement.
\subsection{Cross-modal inference}
The primary objective of this work is to perform cross-modal inference: sampling from $q(Z|X_i)$ allows: \textbf{1.} generative modeling by decoding $p(X_j|Z)$ for $i\neq j$, and \textbf{2.} an estimate of $p(c|Z)$ via Eqn. \eqref{eq:gamma_posterior}. Unfortunately, sampling from the unimodal encoders $q(Z|X_i)$ provides poor embeddings far from the multimodal embedding $q(Z|X_1,...,X_d)$. To remedy this, we introduce a second set of unimodal encoders 
$\tilde{q}(Z|X_i) \sim \mathcal{N}(Z; \tilde{\mu}_i,\tilde{\sigma}_i^2 \mathbf{I})$ with identical architecture to $q(Z|X_i)$. After the multimodal network is trained, we minimize the KL-divergence between $\tilde{q}(Z|X_i)$ and ${q}(Z|X_1,...,X_D)$ so that the unimodal embeddings reproduce the multimodal one. In this sense, the unsupervised multimodal training provides labels which allows supervised training of unimodal embeddings. The KL loss admits the closed form expression
\begin{equation}
    \mathcal{L}_i = \frac12 \sum_{d,k} \left[ \log \frac{\tilde{\sigma}^2_{i,kd}}{\sigma^2_{kd}} - l + \frac{\sigma^2_{kd}}{\tilde{\sigma}^2_{i,kd}} + \frac{\left(\tilde{\mu}_{i,kd} - \mu_{kd}\right)^2}{\tilde{\sigma}^2_{i,kd}}\right],
\end{equation}
which may be sequentially optimized with Adam for each modality $i$ after the multimodal model has been fit. An additional possibility not pursued here is to perform a Bayesian estimate to identify either the cluster most likely to generate the data
\begin{equation}
    p(c|X_i) = \frac{p(X_i|c)p(c)}{\sum_{c'}p(X_i|c')p(c')}\,\,\text{ for }\,\,X_i\in\mathcal{M}_{S},
\end{equation}
or the cluster centroid most likely to have generated the data
\begin{equation}
    p(\mu_c|X_i) = \frac{p(X_i|\mu_c)p(\mu_c)}{\sum_{c'}p(X_i|\mu_{c'})p(\mu_{c'})}\text{ for }\,\,X_i\in\mathcal{M}_{DD}.
\end{equation}

\begin{algorithm}[tb]
   \caption{Training with streaming EM for cluster centers}
   \label{alg:training}
\begin{algorithmic}
   \STATE {\bfseries Input:} data $\mathbf{x}=\{X_1,...,X_d\}$ in batches $\mathcal{B}$
   \FOR{$i=1$ {\bfseries to} $N_{epochs}$}
   \STATE Calculate $\gamma = p(c|\mathbf{x})$ for all $\mathbf{x}$
   \STATE Let $W^{new} = 0$, $M = 0$ 
   \FOR{$\mathbf{b} \in \mathcal{B}$}
   \STATE $W^{old} \gets W^{new}$
   \STATE $W^{new} \gets W^{new} + \sum_{\mathbf{x} \in \mathbf{b}} \gamma_{c}(\mathbf{x})$
   \STATE $M \gets W^{old} M + \frac{\gamma_c\mu}{W^{new}}$
   \ENDFOR
   \STATE $\mu_c \gets M$
   \FOR{$\mathbf{b} \in \mathcal{B}$}
   \STATE Calculate Adam update on ELBO
   \ENDFOR
   \ENDFOR
\end{algorithmic}
\end{algorithm}

\section{Experiments}
\begin{figure}[h]
  \includegraphics[width=0.5\textwidth]{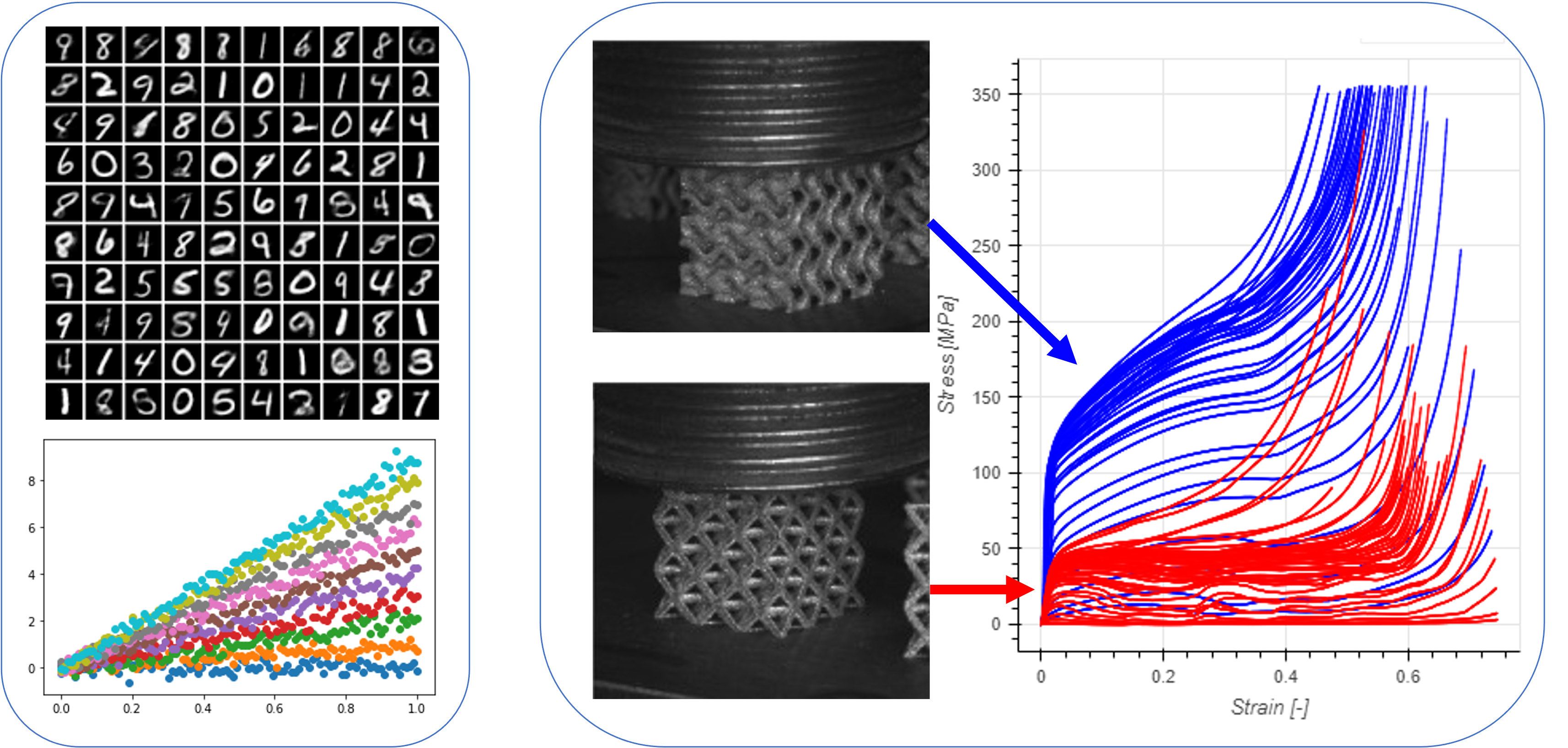}
  \caption{Experimental setup for unsupervised multimodal MNIST \textit{(left)} and multimodal mechanical test \textit{(right)}. For MNIST, we replace the labels on digits $c\in\left\{0,...,9\right\}$ by a sample of the function $X_2 = ct + \epsilon$, for $t \in [0,1]$ and Gaussian noise $\epsilon$; c is therefore encoded either as an image in $X_1$ or as the slope of a 1D function in $X_2$. For the mechanical problem, a high-throughput compression test is performed on two populations of additively manufactured lattices corresponding to gyroid and octet microstructure \cite{garland2020deep}. We supplement the resulting stress-strain curves $X_2$ with images of the microstructure $X_1$ to obtain low- and high-throughput modalities, respectively.}
  \label{fig:expsetup}
\end{figure}

Hyper-parameters for both training and architecture were selected using the Weights and Biases experiment tracking tool \cite{wandb} and were based on the MNIST dataset \cite{lecun1998gradient}
 using a 90/10 train/validation split of the training data. MNIST results are reported on the standard 10,000 held out test examples \cite{lecun2010mnist}. All parameters used in the study, code and data to reproduce experiments, and discussion of hardware and training time for each experiment, may be found in Appendix \ref{app_archhyp}. We consider a non-physical and physical pair of experiments described in Figure \ref{fig:expsetup}. 

\begin{figure}[h]
  \includegraphics[width=0.22\textwidth]{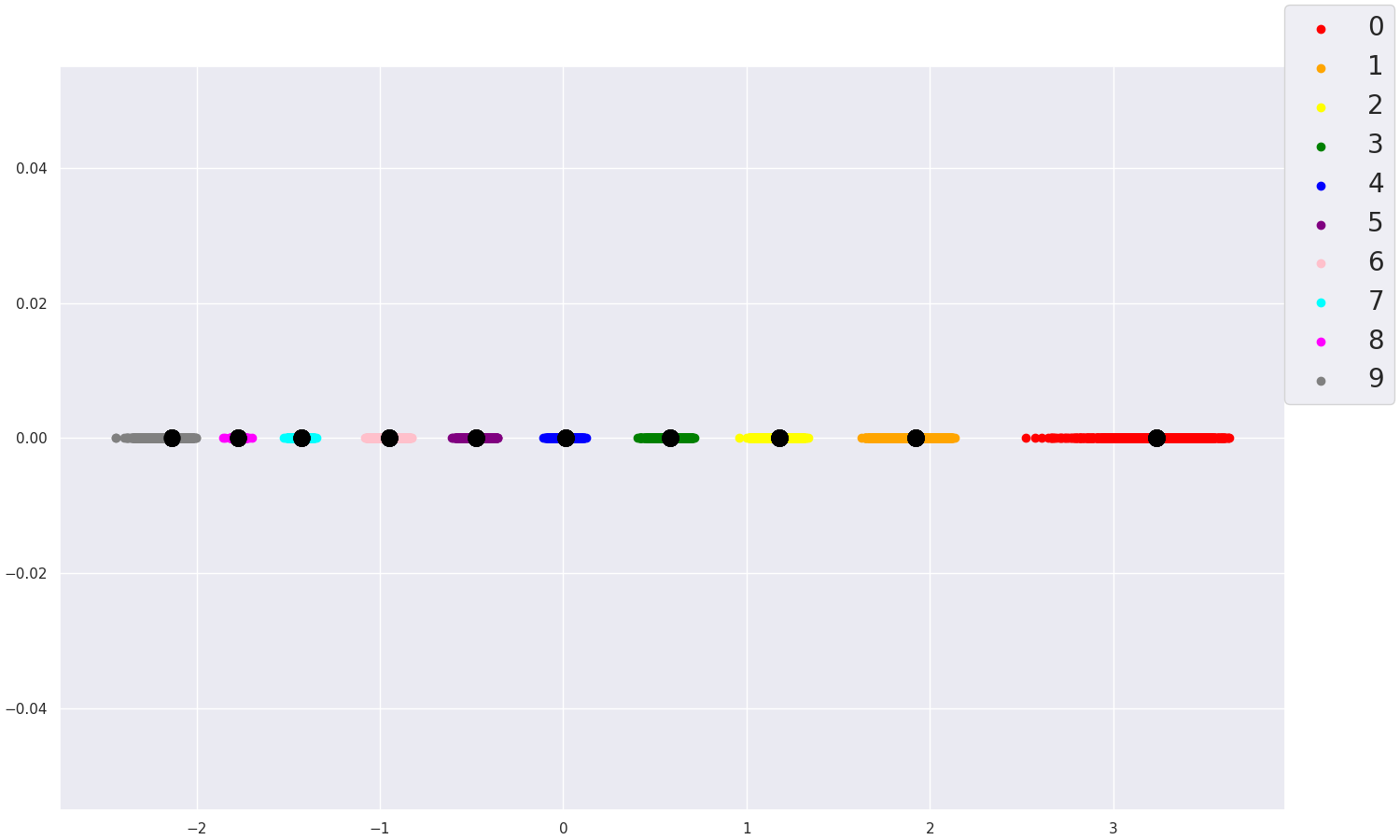}
  \includegraphics[width=0.22\textwidth]{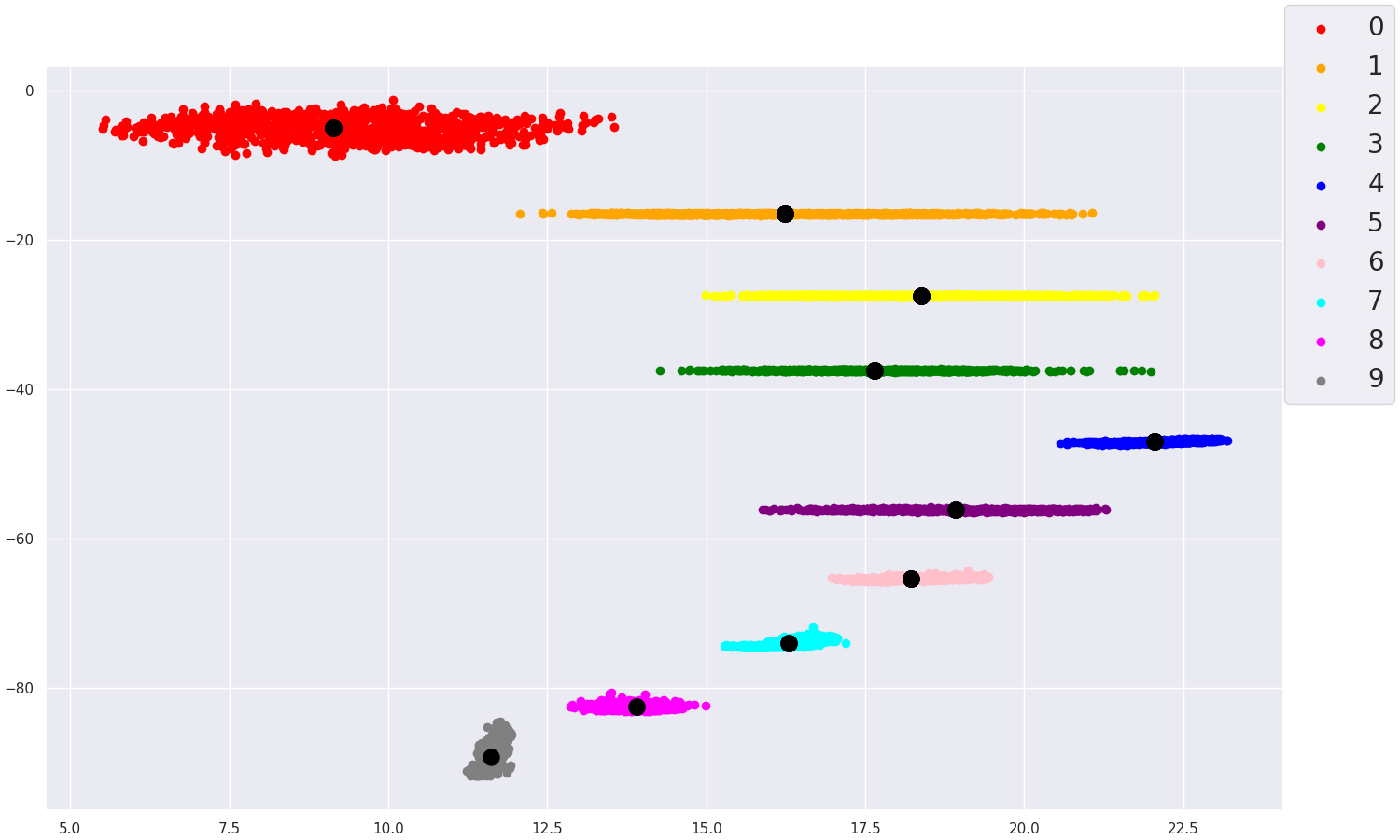}
  \\
  \includegraphics[width=0.22\textwidth]{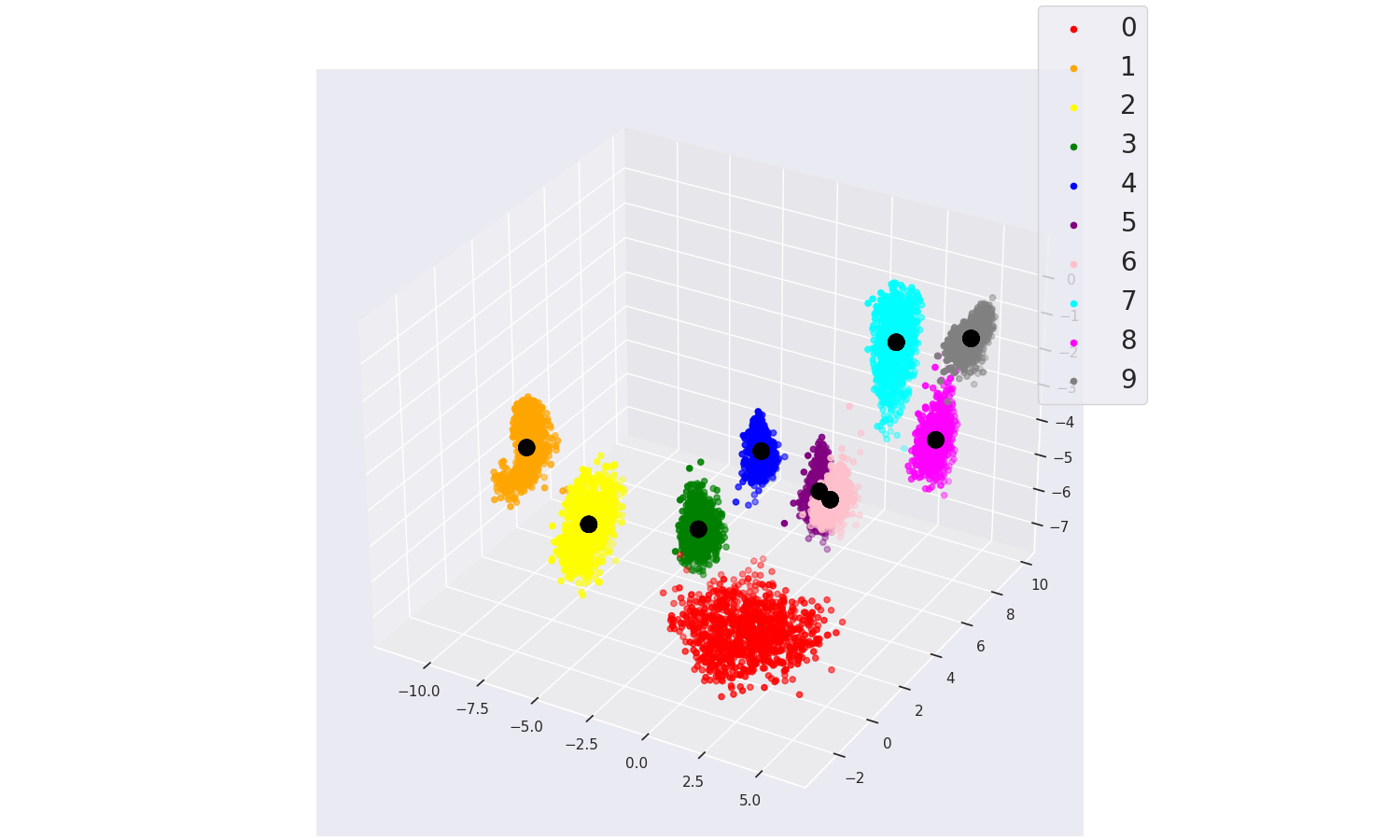}
  \includegraphics[width=0.22\textwidth]{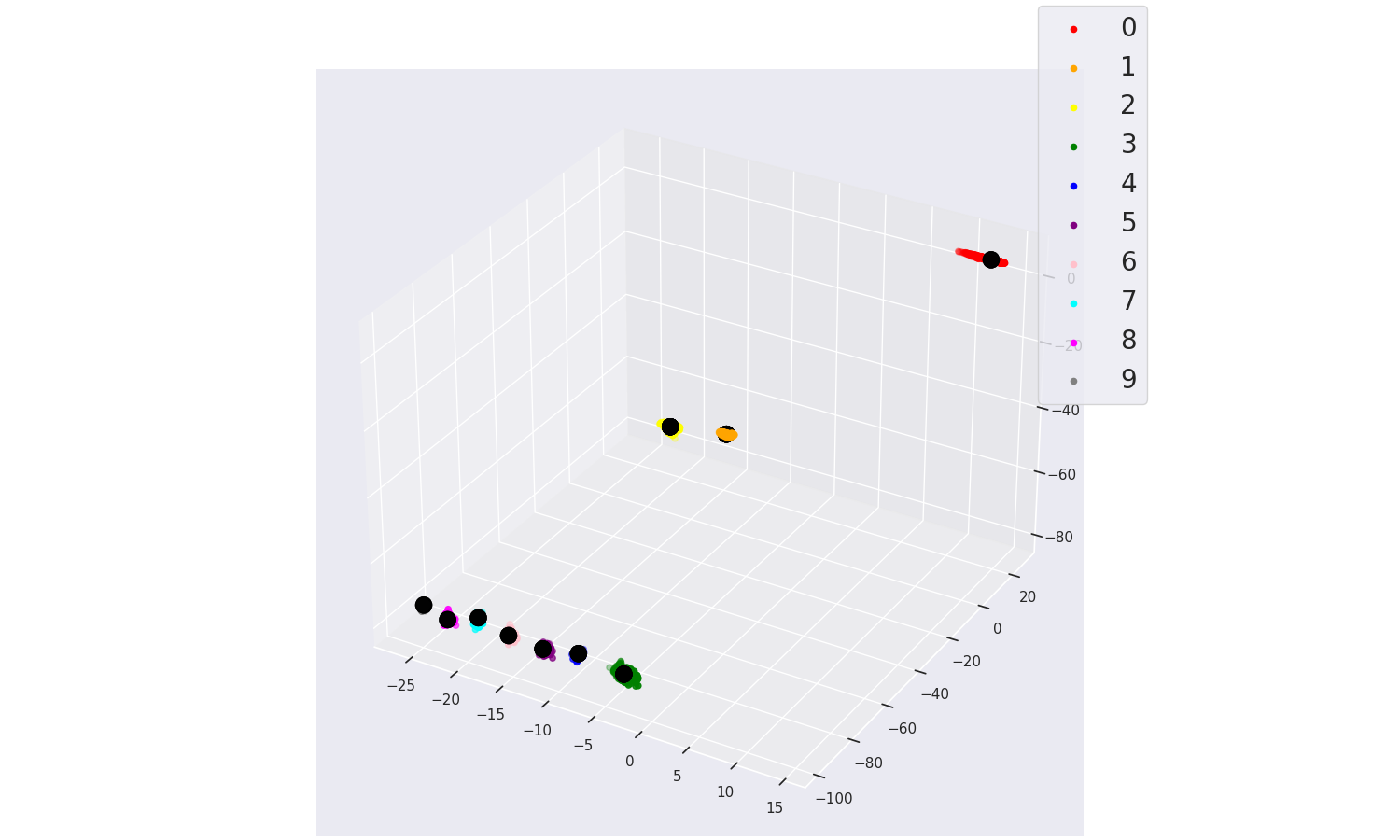}
      \begin{tabular}{c|c|c|c}
        Encoding Dim. & Mean Acc. & Min Acc. & Max Acc. \\
         1 & 0.793 +/- 0.147 & 0.606 & 0.999\\
         2 & 0.854 +/- 0.121 & 0.617 & \textbf{1.00} \\
         3 & \textbf{0.967 +/- 0.047} & \textbf{0.894} & \textbf{1.00} \\
         4 & 0.921 +/- 0.614 & 0.810 & \textbf{1.00} \\
    \end{tabular}
  \caption{MNIST clusters and resulting accuracies for multimodal clean dataset, varying encoding dimension of size 1, 2, 3, and 4 and selecting model from hyperparameter training with lowest validation loss. Test accuracy achieved by clusters shown: 99.9\%, 100.0\%, 99.9\%, and 100.0\%, respectively. Statistics correspond to 10 runs using optimal hyperparameters. 3 of the 4 encoding dimensions from the 4D model are shown.}
  \label{fig:mnistclusters}
\end{figure}

\begin{table*}[]
    \centering
  
       \begin{tabular}{|c|c|c|c|c|c|c|c|}
        \hline
        Method &  CNN SotA$^*$ & VAE+GMM$^\dagger$ & DEC$^\dagger$ &
        VaDE$^\dagger$ & GMVAE$^{\dagger\dagger}$ & GMVAE$^{\dagger\dagger}$\\
        \hline
        Notes & Supervised & & & & 10 clusters & 16 clusters \\
        \hline
        Accuracy (max) & 99.91&  72.94 & 84.30 & 94.46 & 88.54& 96.92  \\
        Accuracy (mean±stdev) &n/a&  n/a & n/a & n/a & 82.31 (3.75)  & 87.82 (5.33)\\
        \hline
    \end{tabular}
        \begin{tabular}{|c|c|c|c|c|c|c|}
        \hline
        Method &   PIMA & PIMA & PIMA & PIMA & PIMA & PIMA\\
        \hline
        Notes & \begin{tabular}[x]{@{}c@{}}multimodal\\low noise\\10 clusters\end{tabular} & \begin{tabular}[x]{@{}c@{}}multimodal\\high noise\\10 clusters\end{tabular} & \begin{tabular}[x]{@{}c@{}}unimodal $X_1$\\low noise\\10 clusters\end{tabular} & \begin{tabular}[x]{@{}c@{}}unimodal $X_2$\\low noise\\10 clusters\end{tabular} & \begin{tabular}[x]{@{}c@{}}unimodal $X_1$\\high noise\\10 clusters\end{tabular} & \begin{tabular}[x]{@{}c@{}}unimodal $X_2$\\high noise\\10 clusters\end{tabular}\\
        \hline
        Accuracy (max)&  100.0 & 98.53 & 79.62 & 99.98 & 84.85 & 78.33 \\
        Accuracy (mean±stdev) &  96.74 (4.72) & 93.15 (7.07) & - & - & - & -\\
        \hline
    \end{tabular}

    \caption{Unsupervised classification accuracy for MNIST. Results gathered from \cite{an2020ensemble}, \cite{jiang2016variational} and \cite{dilokthanakul2016deep} denoted by $*$,$\dagger$ and $\dagger\dagger$, respectively. If statistics were not provided we assume maximum accuracy was reported. While the data augmentation offered by $X_2$ is not incorporated in comparisons to unimodal unsupervised benchmarks, a comparison to the supervised setting is valid. For all experiments we do not overparameterize and keep clusters equal to the number of digits.}
    \label{tab:mnist_comparison}
\end{table*}

\begin{figure*}[h]
\centering
  \includegraphics[width=0.44\textwidth]{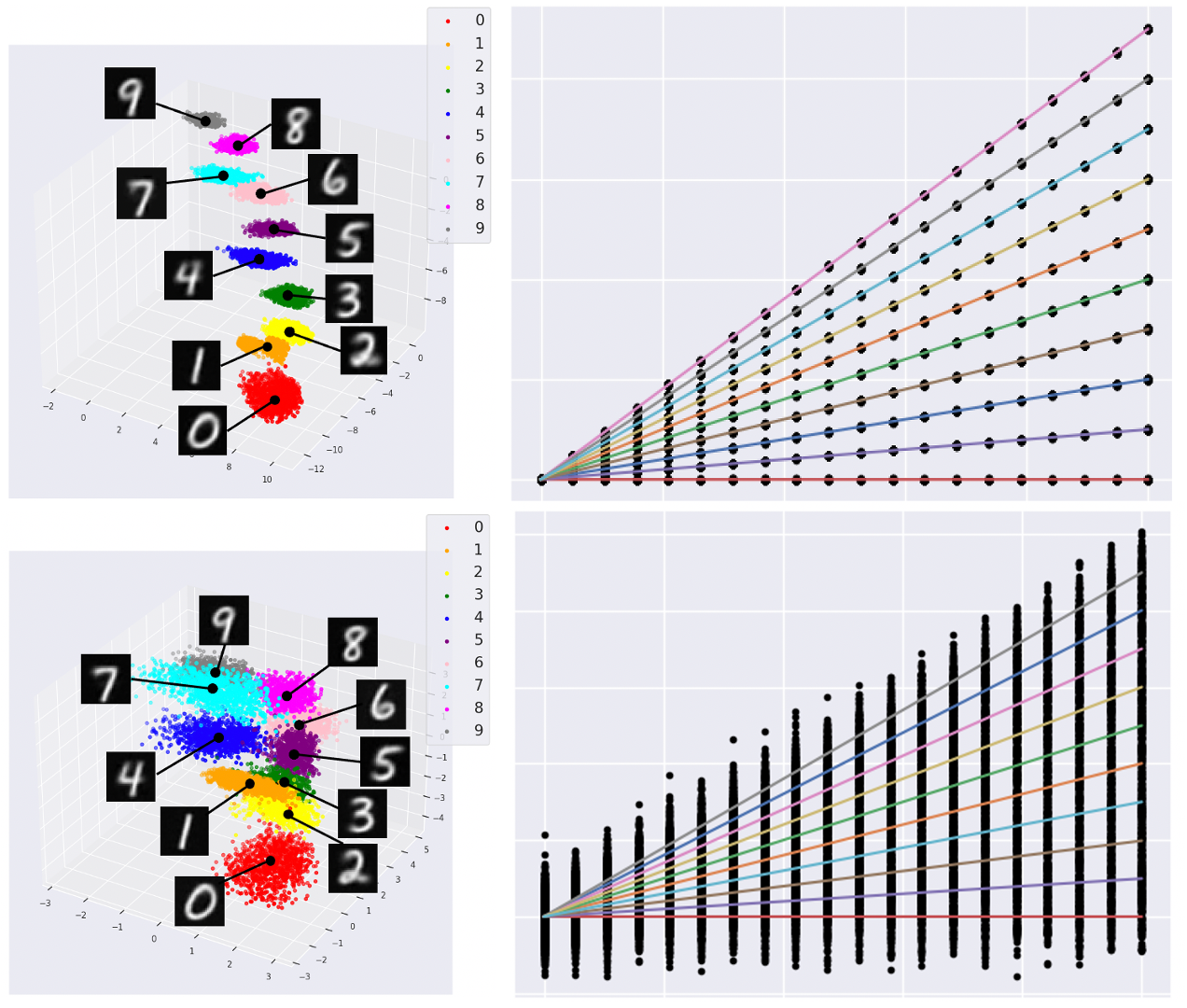}
    \includegraphics[width=0.37\textwidth]{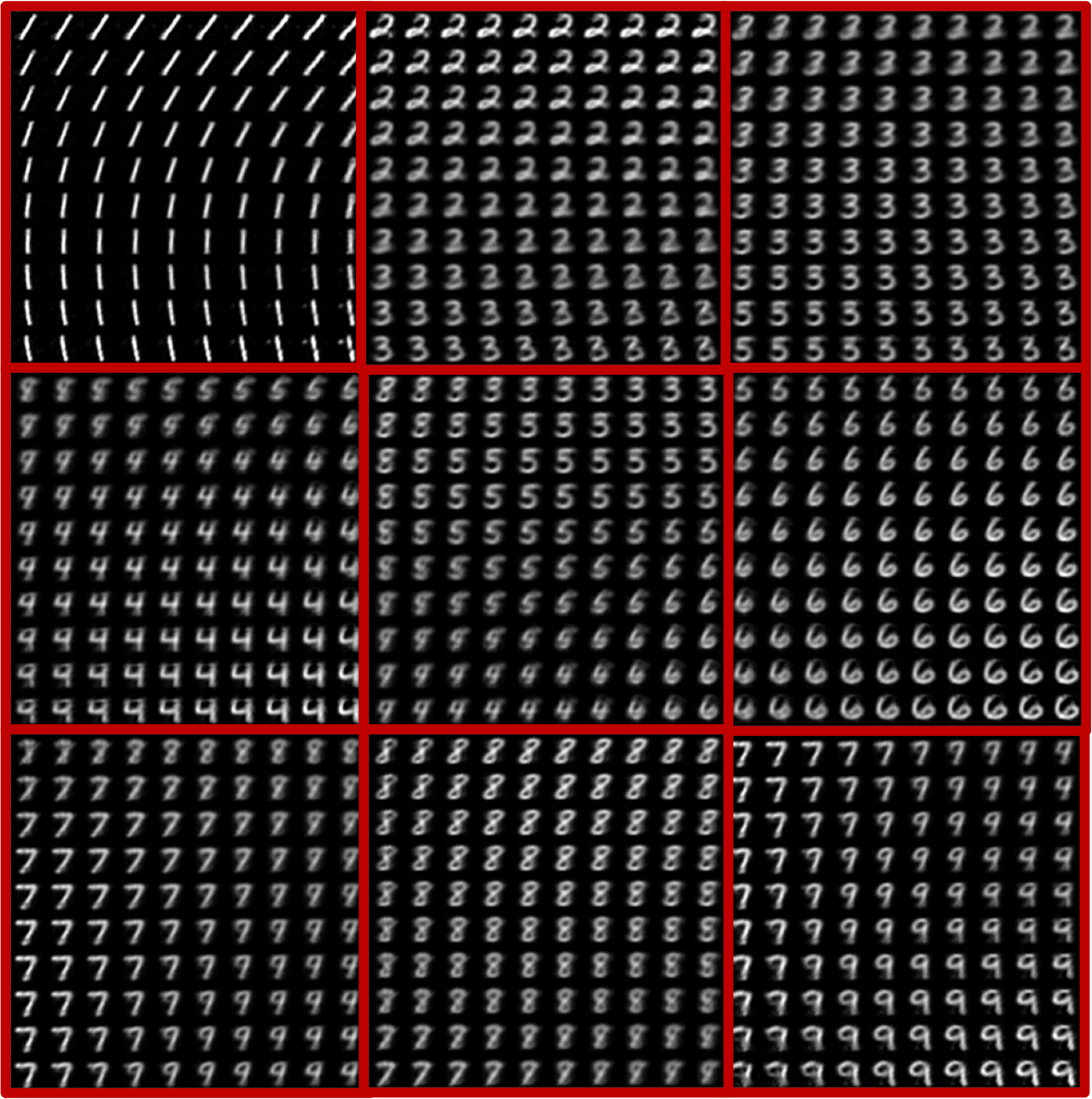}
  \caption{Generative modeling for multimodal MNIST experiment. Clustering in latent space with generative digit images $X_1$ at cluster centroids \textit{(left)} and expert fit to $X_2$ \textit{(center)} for clean \textit{(top)} and noisy \textit{(bottom)} $X_2$ data. In both cases, data is embedded into disentangled clusters adjacent to digits to obtain ordered clusters; for noisy data clusters overlap to reflect class confusion in $X_2$ but preserve ordering. In this noisy case, this is reflected when sampling $X_1$ from $Z_i \in \left\{\mu_c \text{±}  3\sigma_c\right\}_i$ for clusters 1-9 \textit{(right)}: at the periphery of each cluster a digit ±1 is generated, reflecting the fuzzy class boundary of $X_2$.}
  \label{fig:mnistgenerative}
\end{figure*}

\begin{figure}[h]
\centering
\includegraphics[width=0.225\textwidth]{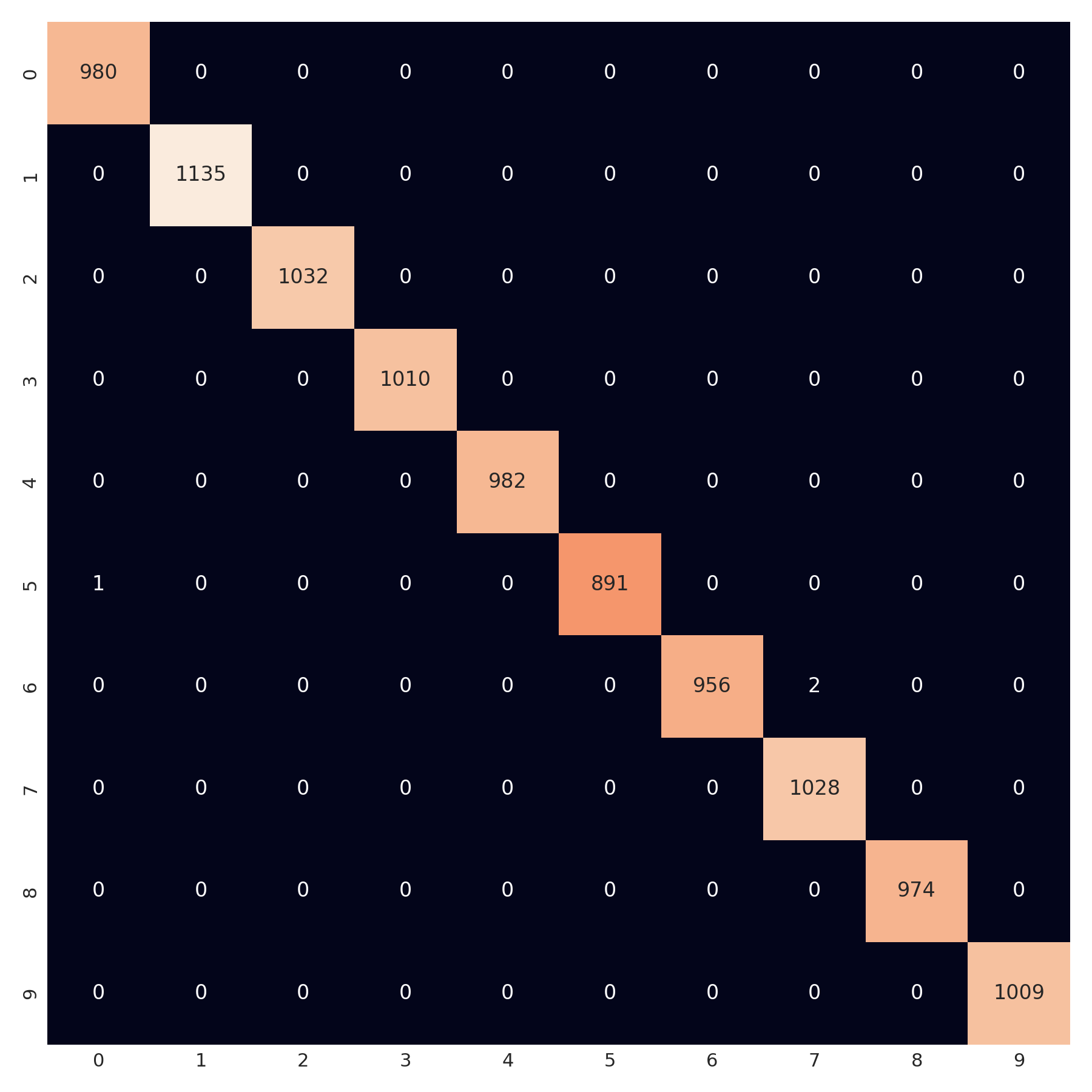}
\includegraphics[width=0.225\textwidth]{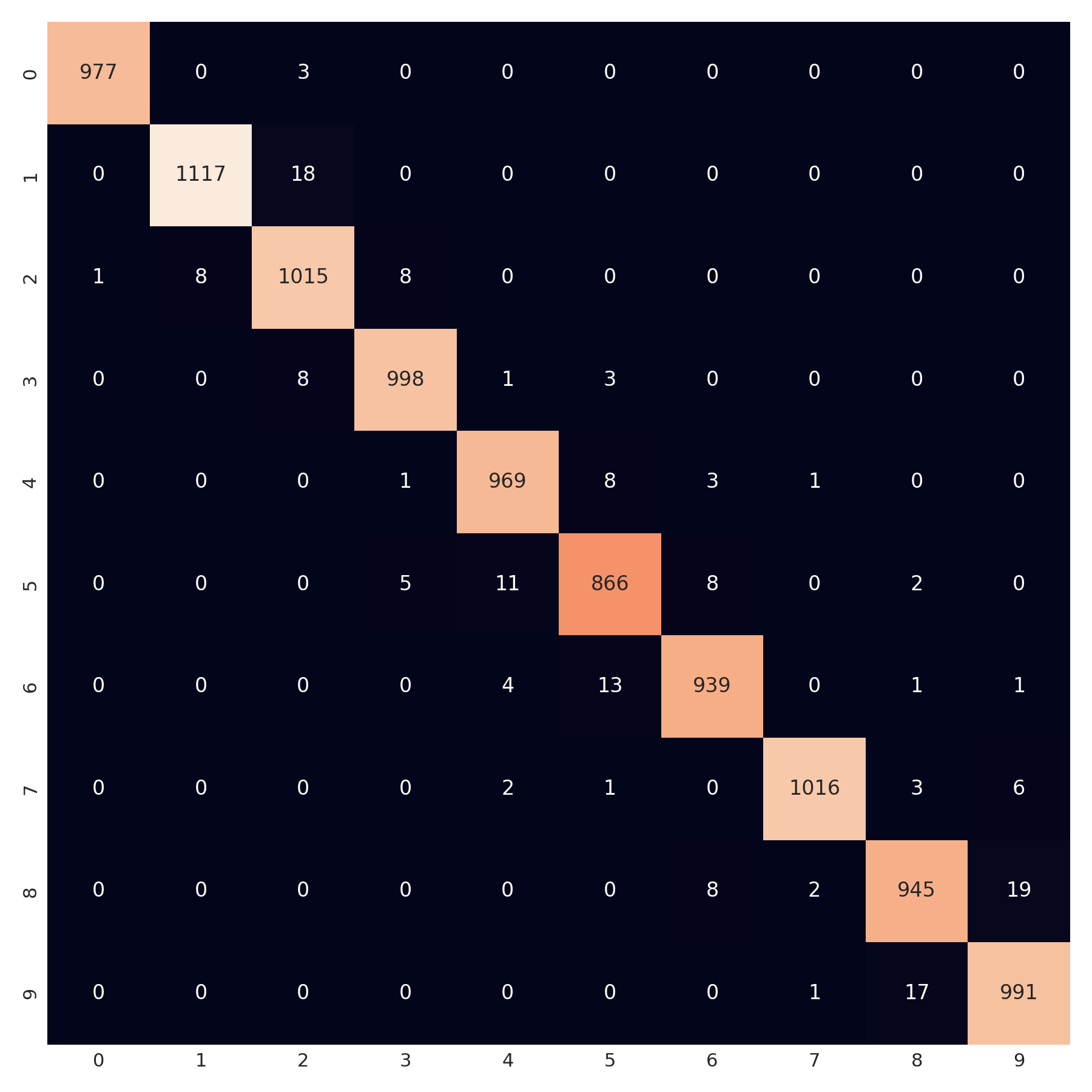}\\
\includegraphics[width=0.225\textwidth]{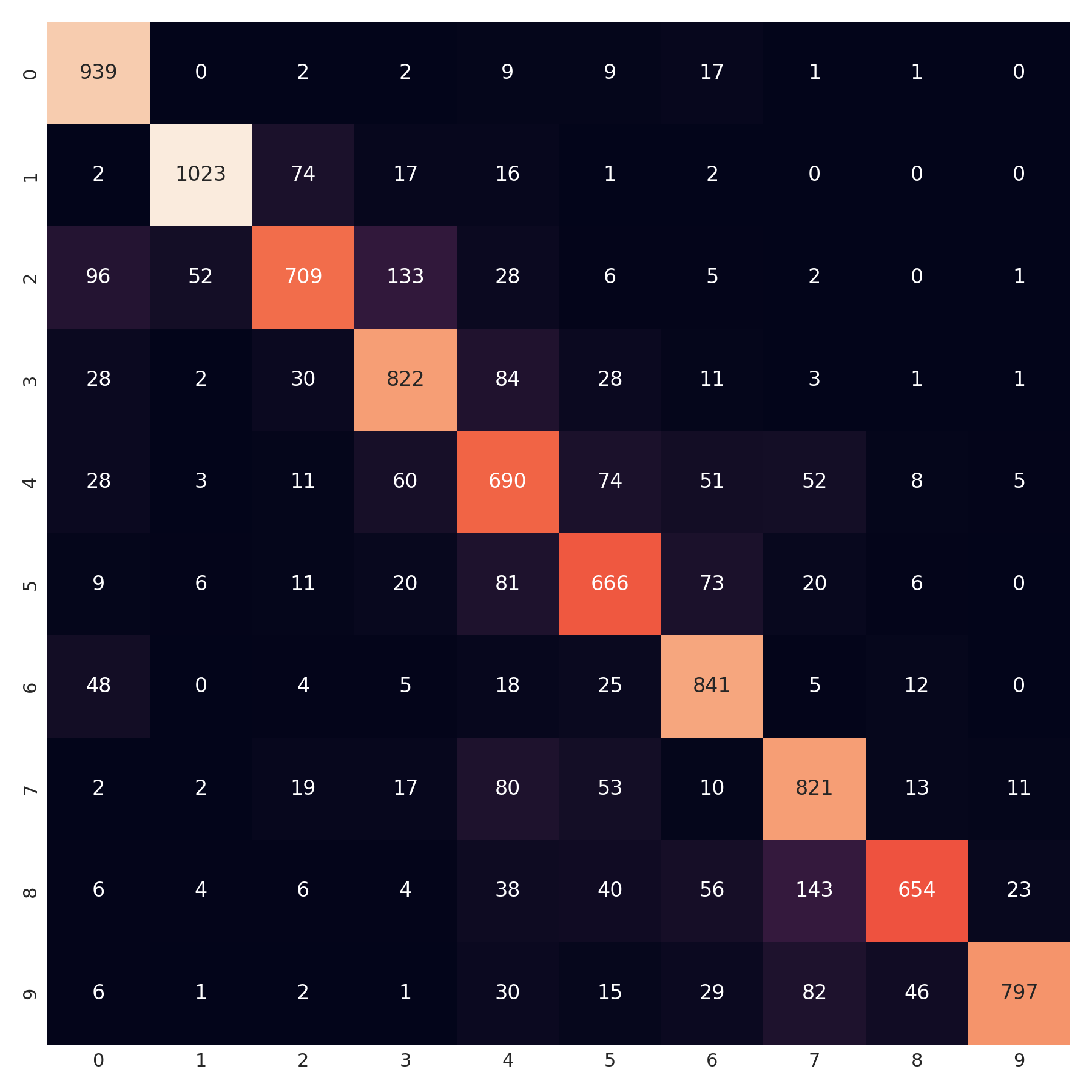}
\includegraphics[width=0.225\textwidth]{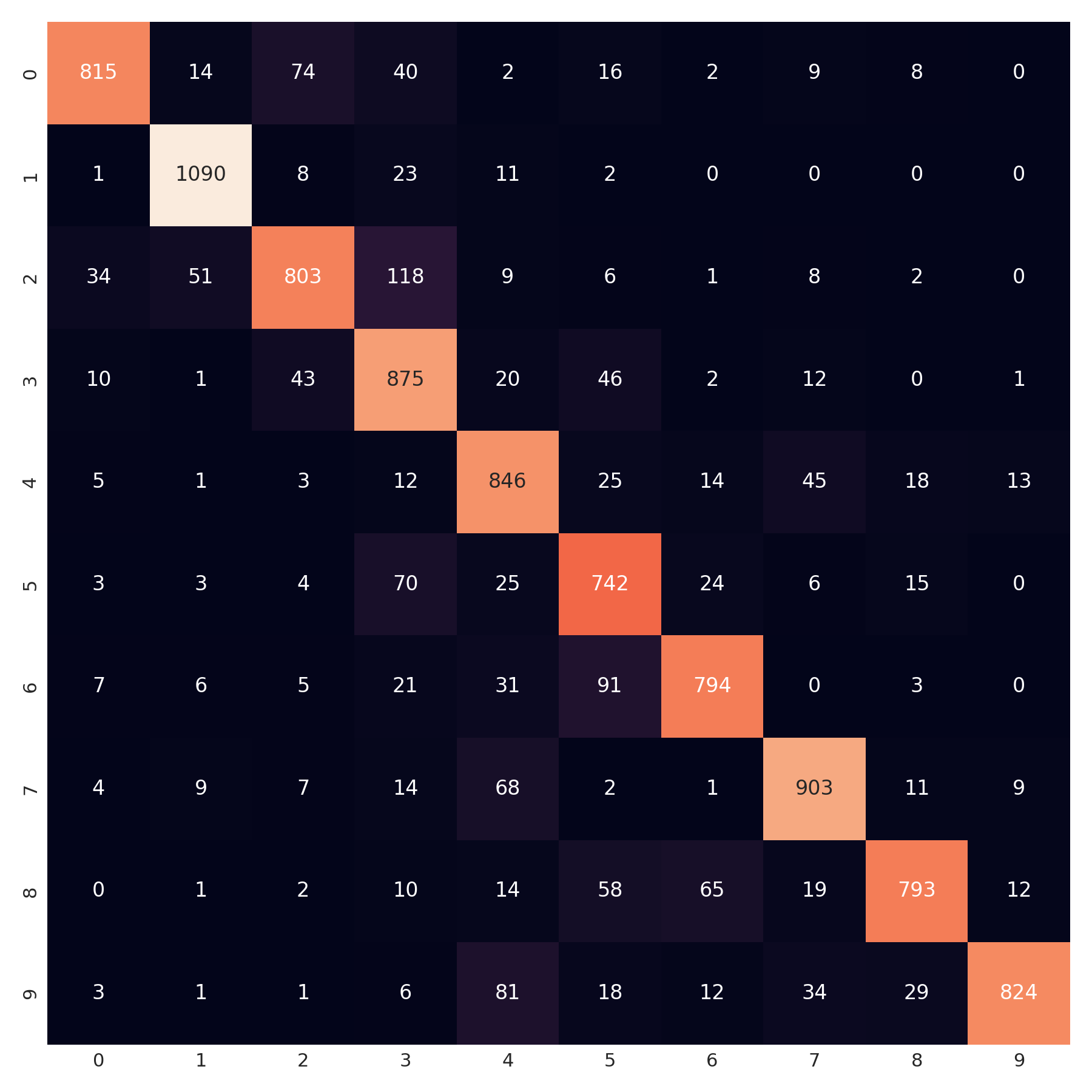}\\
\includegraphics[width=0.225\textwidth]{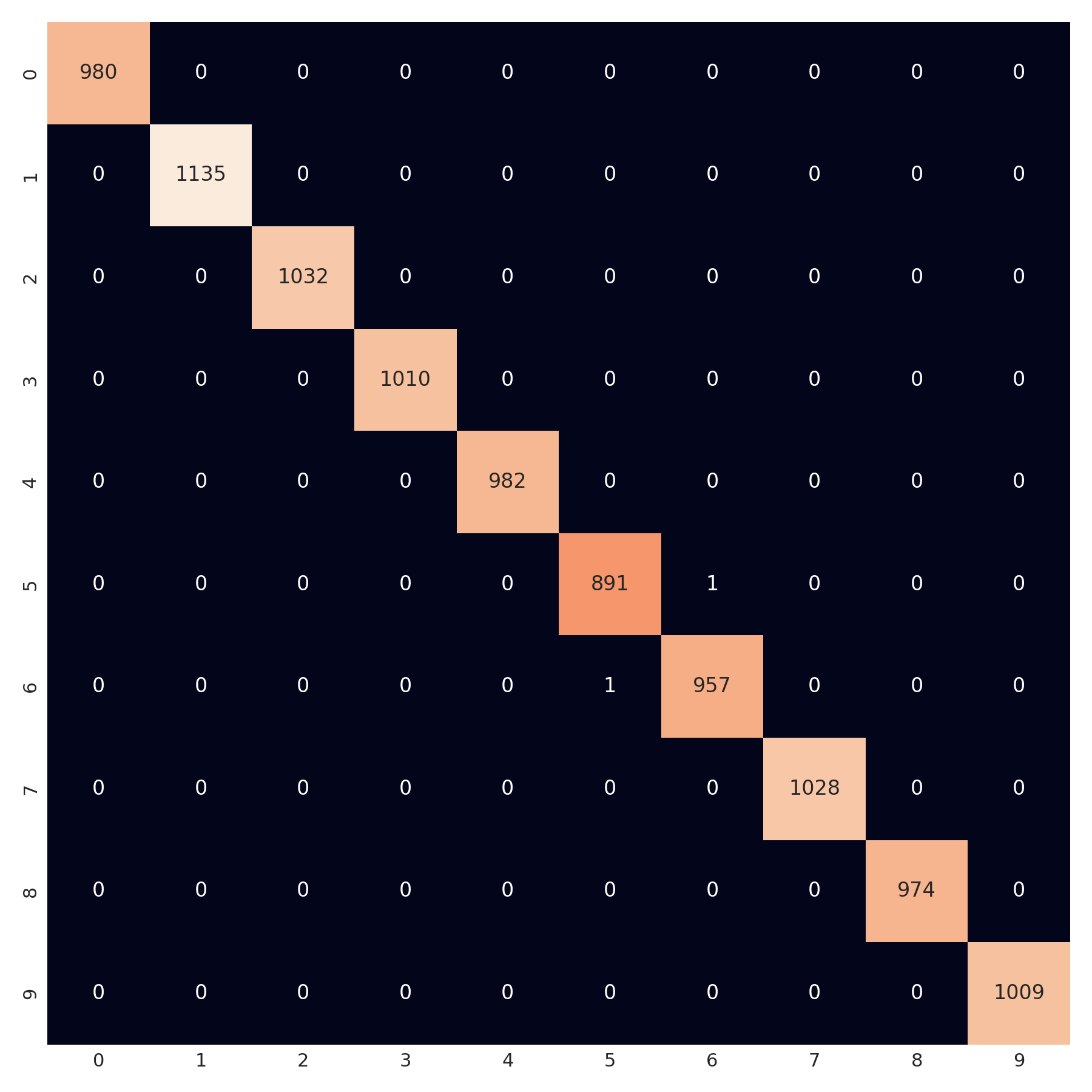}
\includegraphics[width=0.225\textwidth]{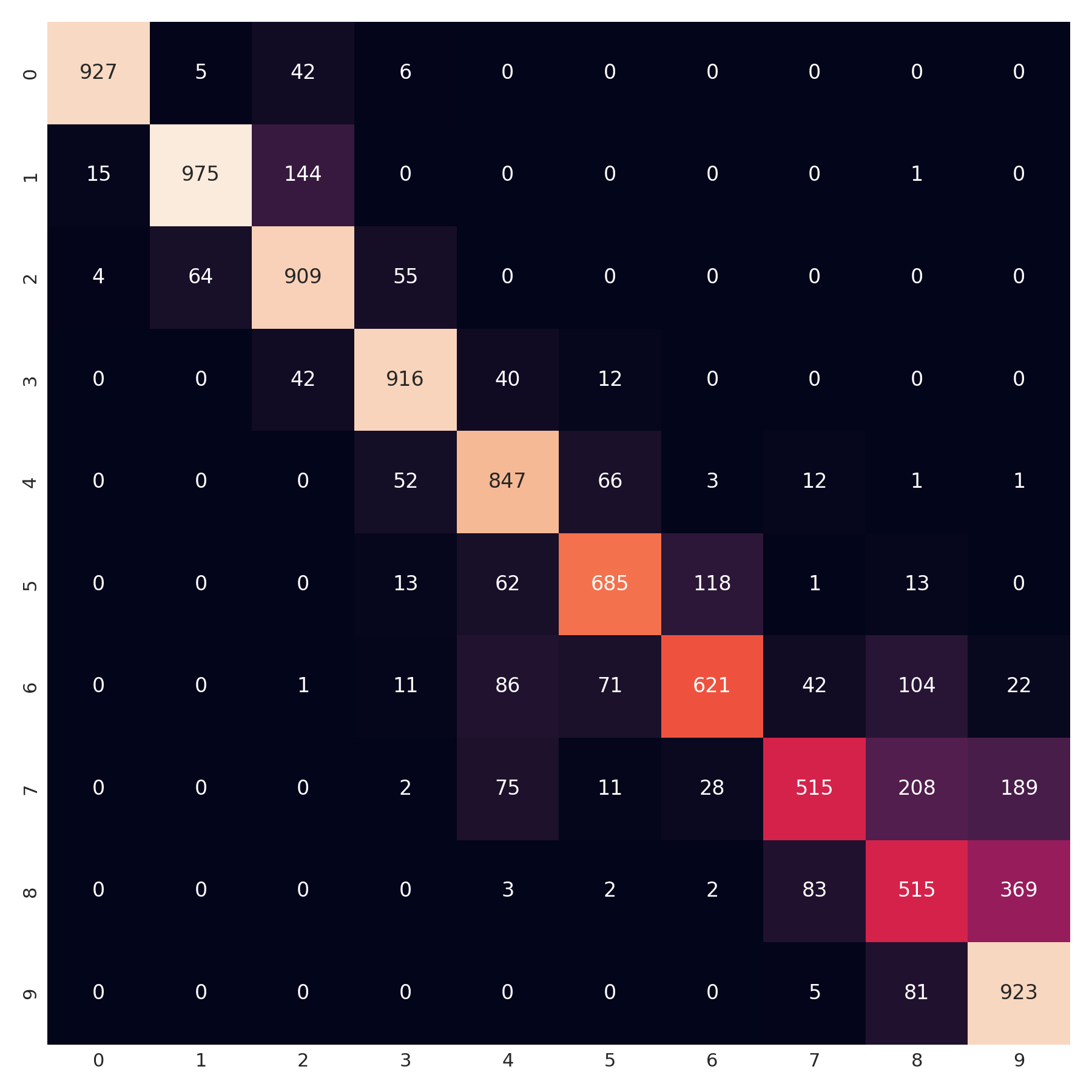}
  \caption{Confusion matrices denoting frequency with which each cluster reproduces labels during unsupervised training. The approximate banding of the matrix illustrates that the sequential embedding of clusters limits misclassified digits to numbers with similar values in $X_2$. \textit{Left/right:} clean/noisy, \textit{top:} Multimodal inference \textit{center/bottom:} inference from $X_1$/$X_2$, respectively.}
  \label{fig:mnistconfuse}
\end{figure}

\subsection{Unsupervised multimodal MNIST}
In the first experiment, we take the traditional MNIST dataset consisting of images $X_1$ and labels $c\in\left\{0,...,9\right\}$ and manufacture a synthetic 1D modality $X_2 = ct + \epsilon$, where $t \in [0,1]$ and $\epsilon \sim \mathcal{N}(0,0.01)$ for a clean dataset or $\epsilon \sim \mathcal{N}(0,0.5)$ for a noisy data set. We adopt the affine expert model $\mathcal{E}(t;\theta_c) = \theta_c t$, and perform unsupervised clustering of the multimodal dataset $(X_1,X_2)$, and additionally perform cross-modal inference. For this artificial problem, the labels are thinly veiled as the slope of second modality, and so we expect that if we successfully perform multimodal inference we should obtain accuracy comparable to a supervised MNIST benchmark.

We define unsupervised clustering accuracy ($acc$) as in \cite{xie2016unsupervised}, \cite{jiang2016variational}:
\begin{equation}
    acc = \underset{m \in M}{\text{max}} \frac{\sum_{i=1}^{N} \mathbbm{1} \{l_i=m(c_i)\}}{N},
\end{equation}
where $N$ is the number of examples, $M$ is the set of all possible mappings from a cluster to a label assignment, $l_i$ is the true label and $c_i$ is the cluster assignment by the model for example $i$.
Calibration of the latent dimension (Figure \ref{fig:mnistclusters}) revealed $l=3$ to be an optimal embedding dimension. In Table \ref{tab:mnist_comparison} we provide a comparison to classification accuracy against state of the art supervised and unsupervised models trained on images only. For multimodal inference with low noise we obtain perfect classification outperforming the current state-of-the-art in supervised classification and outperforming the state-of-the-art in unsupervised learning. 

In addition to observing improved accuracy, we investigate in Figure \ref{fig:mnistgenerative} the disentangled representation of clusters and provide examples of generative models. Surprisingly, the sequential ordering of clusters in $X_2$ induces an sequential embedding of corresponding clusters in latent space. As noise is increased, the data in $X_2$ is sufficiently large to prevent distinct clustering into disentangled classes, however the ordering of digits is roughly preserved. When generative modeling is performed for $X_2$, this is reflected by samples from the tail of the Gaussian mixture which generate images of adjacent digits. For example, the cluster of 2's contains images of 3's  in its periphery. This suggests that the sequential ordering of data in $X_2$ may induce generative models for $X_1$ which reflect cross-modal ordering. 
Confusion matrices for multimodal and cross-modal classification in Figure \ref{fig:mnistconfuse} reveal an approximately banded structure, whereby misclassified modalities primarily occur between adjacent digits. While this single experiment is insufficient to remark on the generality of this result, it suggests the potential for learning generative models of images which reflect information from the expert model, a particularly exciting prospect for scientific datasets.




\subsection{Metal additive lattice fingerprinting}
\begin{figure}[h]
  \includegraphics[width=0.5\textwidth]{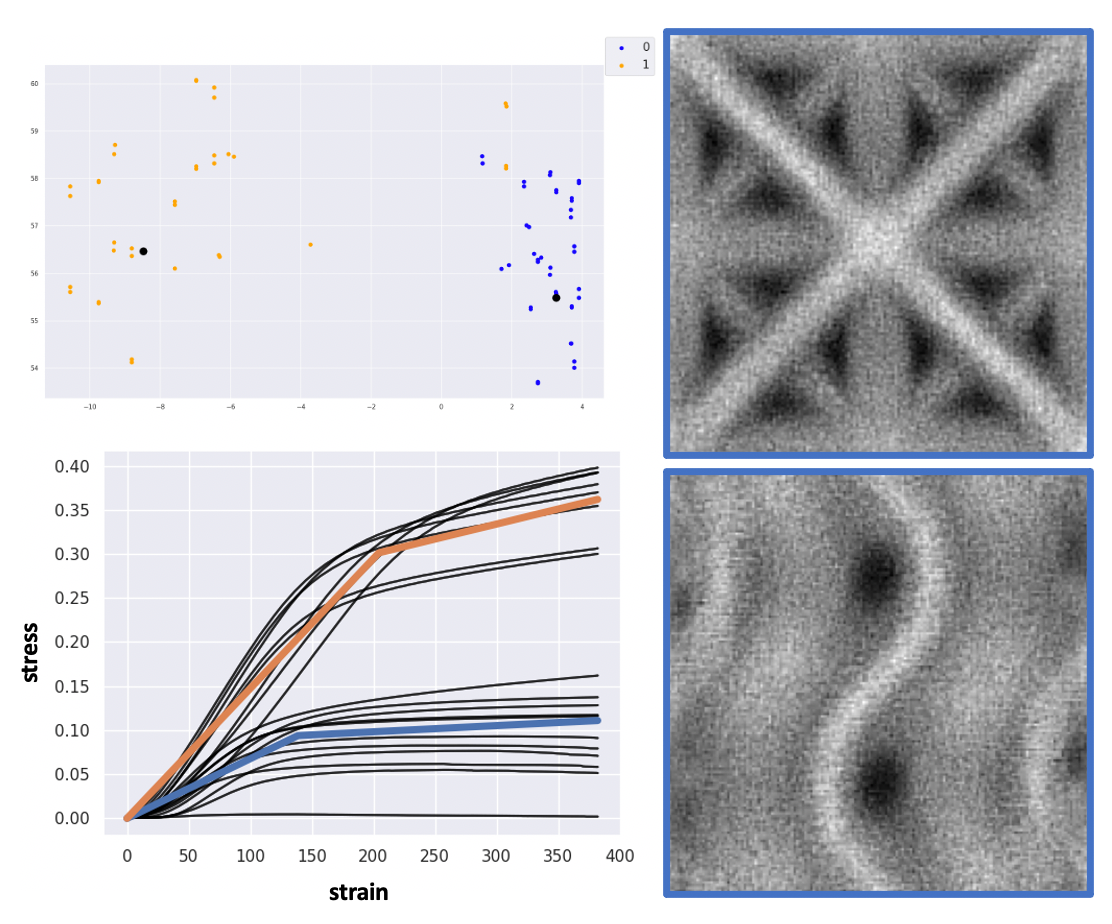}
  \caption{Generative modeling for metal additive dataset. \textit{Top left:} Data is clustered into two gaussian distributions with mean visualized by black dot. Shown here is test data held out during training. \textit{Bottom left:} Stress-strain curves from two populations split naturally between two populations, and linear-strain hardening expert model provides provides best fit (orange and blue lines). \textit{Right:} Generative sampling from the means of each cluster provides a prediction for microstructure associated with either octet \textit{(bottom)} or gyroid \textit{(top)} lattice topologies.}
  \label{fig:latticemicro}
\end{figure}
The lattice dataset consists of $X_1$ images of 3D printed lattices split between two types of printed metamaterials and corresponding $X_2$ stress/strain curves performed in a high-throughput uniaxial compression machine. Even with high-throughput testing, only 91 pairs of images were able to be generated, highlighting the utility of high-throughput photographs $X_1$ as surrogates to infer $X_2$. We select as expert model a linear strain-hardening model partitioning the stress-strain response into two piecewise linear regions \cite{jones2009deformation}. Even a simple model like this succinctly encodes a large number of interpretable quantities of interest: a yield stress, together with elastic and plastic moduli. We gather in Figure \ref{fig:latticemicro} results from generative modeling and include in the appendix additional results demonstrating 94.74\%/94.74\%/94.74\% classification accuracy of the two clusters for multimodal/$X_1$/$X_2$-cross-modal inferences, respectively.

\section{Conclusions and future work}

The present approach provides an abstract variational inference for discovering fingerprints in an unsupervised manner while incorporating physical model biases. This framework is widely applicable to a range of scientific disciplines where detection of fingerprints are crucial for tasks ranging from predicting and attributing climate change to designing biochemical pathways at a molecular level. In addition to the application focus on fingerprint generation, this framework may be used for a variety of general purpose downstream tasks based on multimodal processing of scientific data. For this work we have focused on a simple MNIST example to probe dynamics for an easily replicable and understandable dataset, along with a simple high-throughput manufacturing example to illustrate feasibility for facilitating high-throughput experimentation. 

In future work, we will employ more sophisticated physics-informed surrogates as expert models for processes in additive manufacturing and semiconductor device design bridging multiscale and multifidelity information from data sources corresponding to both physical experiment (transmission electron microscopy, atom probe tomography, micro x-ray computerized tomography, or synchotron x-ray diffraction) and high-fidelity simulations (quantum density functional theory, molecular dynamics, crystal plasticity, and continuum mechanics). To date, these costly but rich sources of information are antagonistic to high-throughput testing and simulation. This framework provides an exciting platform for discovering data-driven scientific fingerprints which may be combined with advances in automated experimentation to accelerate scientific discovery.



\section*{Software and Data}

Pending acceptance, all data and code used to generate results will be hosted on a Github page. For now, we include a subset of the code to reproduce the MNIST examples through the anonymous Github\footnote{\url{https://anonymous.4open.science/r/PIMA-5D6B/}}; unfortunately anonymous github does not offer enough storage to host the accompanying dataset.

\section*{Acknowledgements}
The authors thank Warren Davis, Anthony Garland and Lekha Patel for providing guidance on the variational inference framework and review of the manuscript and  Kat Reiner and Greg Geller for providing computing support. All authors acknowledge funding under the Beyond Fingerprinting Sandia Grand Challenge Laboratory Directed Research and Developement program. N.~Trask acknowledges funding under the Collaboratory on Mathematics and Physics-Informed Learning Machines for Multiscale and Multiphysics Problems (PhILMs) project funded by DOE Office of Science (Grant number DE-SC001924) and the DOE Early Career program. Sandia National Laboratories is a multi-mission laboratory managed and operated by National Technology and Engineering Solutions of Sandia, LLC., a wholly owned subsidiary of Honeywell International, Inc., for the U.S. Department of Energy’s National Nuclear Security Administration under contract DE-NA0003525. This paper describes objective technical results and analysis.  Any subjective views or opinions that might be expressed in the paper do not necessarily represent the views of the U.S. Department of Energy or the United States Government. SAND number: SAND2022-1159 O
\bibliography{icmlBFPpaper}
\bibliographystyle{icml2022}

\clearpage
\appendix
\onecolumn
\section{Architecture, hyperparameters, and implementation}\label{app_archhyp}
\textbf{Model Architectures}

We employ relatively small convolutional architectures to serve as encoders for both modalities.  The image modality encoder consists of 2 2D convolutional layers with 32 and 64 channels respectively, each with 3x3 kernels. We apply the exponential linear unit (ELU) activation function as well as  batch normalization after each convolutional layer, then pass the output to a fully connected layer of size $encoding_dim \times 2$ to enable an embedding into a representation of the mean and standard deviation of the input in the latent space.  For the 1D modality encoder, in place of 2D convolutional layers, we use 1D convolutions with 8 and 16 channels in the respective layers, but with an otherwise identical architecture.  

The image decoder begins with a fully connected layer of appropriate size to be reshaped into 32 channels of 2D arrays, with each dimension having a length $\frac{1}{4}$ of the length of the number of pixels per side of the original image. The reshaped output of the initial dense layer is passed into a series of 3 deconvolutional layers with 64, 32, and 1 channel respectively, each with a kernel size of 3. The first 2 deconvolutional layers use a stride of 3 and a Rectified Linear Unit (ReLU) activation function, and the final deconvolutional layer uses a stride of 1. Zero padding is used to retain the input shape while traversing these layers.

\textbf{Hyperparameters}

We used the MNIST dataset along with the Weights and Biases tool \cite{wandb} to perform a hyperparameter search over learning rates and encoding dimension size.  For the multimodal models, we found that a learning rate of $1.97e-5$ gave the best result as measured by validation loss. For the MNIST dataset, an encoding dimension of size 3 gave the most consistent results, and we used a 2D encoding dimension for the lattice dataset. For the unimodal models, we found that a learning rate of $4.398e-5$ for the image modality and a learning rate of $4.398e-3$ for the 1D modality supported learning, but we leave a rigorous hyperparameter tuning for unimodal models for future work.

\textbf{Implementation details}

Each of the input data modalities was normalized to have values in $[0,1]$.  The 1D data was sampled to generate an array of length 20 for MNIST and length 100 for the lattice stress-strain data.  The lattice images were cropped and subsampled into quadrants resulting in images of dimension 152x152.  We further augmented the dataset by flipping
the images along each axis.   

To train the MNIST multimodal models, we used 10\% of the standard MNIST train set for validation, and selected the model with the lowest validation loss.  We did not observe any indication of overfitting the multimodal model to the training data. All results reported are on the held out test dataset.  For the MNIST unimodal models, we again did not observe overfitting when measuring model performance by accuracy to the cluster labels generated from the trained multimodal model, and we report results on the test set from the model resulting from the final training epoch.

We split the lattice dataset into a train/test 80/20 split.  Since we did not observe overfitting during multimodal training, we selected the model with the lowest training loss and applied that model to the held out test set, and we report those results. We select the model resulting from the final training epoch for the unimodal tasks.

Our models are implemented in Python using Tensorflow \cite{abadi2016tensorflow}, and we leverage the Scikit-learn library \cite{pedregosa2011scikit} for data preparation and accuracy metrics, Scipy \cite{virtanen2020scipy} for data preparation and the \texttt{linear\_sum\_assignment} implementation of the Hungarian method \cite{kuhn1955hungarian} for efficient computation of the unsupervised cluster accuracy. We visualize our results using the Matplotlib \cite{hunter2007matplotlib} and Seaborn \cite{Waskom2021} libraries.  Training was performed on NVIDIA DGX-1 and DGX-2 machines with each run executed on 1 GPU.  A combination of P100 and V100 GPUs were used in this work.  Training runs took on average approximately (machine dependent) 10-14 hours with our longest run on lattice data taking approximately one day. We made no attempt to optimize parallel training to improve run time or training efficiency.  

\section{Derivation of ELBO}\label{app_elbo}
To derive a closed form expression for the single sample ELBO
\begin{equation}
    \mathcal{L}_d = \mathbb{E}_{q(Z,c|X_1,...,X_D)}\left[\log \frac{p(X_1,...,X_D,Z)}{q(Z,c|X_1,...,X_D)} \right]
\end{equation}
we apply the separability assumptions
\begin{align}
    p(X_1,...,X_D,Z,c) = \left(\Pi_{i=1}^D p(X_i|Z,c)\right) p(Z|c) p(c), \\
    q(Z,c|X_1,...,X_D) = q(Z|X_1,...,X_D)q(c|X_1,...,X_D). \label{eq:q_z_c}
\end{align}
providing the additive decomposition
\begin{equation}
    \mathcal{L} = \mathbb{E}_{q(Z,c|X_1,...,X_D)}\left[\log p(X_1,...,X_D,Z)\right] -  \mathbb{E}_{q(Z,c|X_1,...,X_D)}\left[\log q(Z,c|X_1,...,X_D)\right]
    \end{equation}
    \begin{equation*}
    = \sum_{i=1}^D \mathbb{E}_{q(Z,c|X_1,...,X_D)}\left[\log p(X_i|Z,c)\right] 
    +\mathbb{E}_{q(Z,c|X_1,...,X_D)}\left[\log p(Z|c)\right] 
    \end{equation*}
    \begin{equation*}
        +\mathbb{E}_{q(Z,c|X_1,...,X_D)}\left[\log p(c)\right] -\mathbb{E}_{q(Z,c|X_1,...,X_D)}\left[\log q(Z|X_1,...,X_D)\right] 
        \end{equation*}
    \begin{equation*}
        -\mathbb{E}_{q(Z,c|X_1,...,X_D)}\left[\log q(c|X_1,...,X_D)\right].
    \end{equation*}
For convenience we denote $\mathbb{E}_{q(Z,c|X_1,...,X_D)}=\mathbb{E}_q$. The separability assumptions therefore decompose the ELBO into constituent expectations of the form
\begin{equation}
    \mathbb{E}_{q}\left[\log f(Z,c)\right] = \sum_c  \int_{\mathbb{R}^l} f(Z,c) \log g(Z,c)\, dZ
\end{equation}
which may be integrated exactly for the Gaussian/categorical $f$ and $g$ appearing in the ELBO. The only term which may not be immediately computed is $\mathbb{E}_{q}\left[\log q(c|X_1,...,X_D)\right]$. The lack of a reparameterization trick for the categorical distribution precludes backpropagation into the encoder, forcing us to consider an encoder which only provides predictions for $Z$. While there are options to use e.g. a regularized Gumbel-softmax approximation to the categorical distribution \cite{jang2016categorical}, we would lose the tractability of the closed form expression for the ELBO. Instead we follow \cite{jiang2016variational} and approximate $q(c|X_1,...,X_D) = p(c|Z)$ using the following justification. 

Rewriting the ELBO
\begin{equation}
    \mathcal{L}_d = \mathbb{E}_{q(Z,c|X_1,...,X_D)}\left[\log \frac{p(X_1,...,X_D,Z)}{q(Z,c|X_1,...,X_D)} \right]
\end{equation}
\begin{equation*}
    = \mathbb{E}_{q(Z,c|X_1,...,X_D)}\left[\log \frac{p(X_1,...,X_D|Z)p(Z)}{q(Z|X_1,...,X_D)} + \log \frac{p(c|Z)}{q(c|X_1,...,X_D)} \right]
\end{equation*}
\begin{equation*}
    = \int_{\mathbb{R}^l} \log \frac{p(X_1,...,X_D|Z)p(Z)}{q(Z|X_1,...,X_D)} dZ + D_{KL}(q(c|X_1,...,X_D)||p(c|Z)),
\end{equation*}
we seek extremal points with respect to $c$. The first term is independent of $c$, and the positive second term takes zero value when $q(c|X_1,...,X_D) = p(c|Z)$, providing the desired maximum. We caution however that this holds only at local minima of the loss landscape, but empirically has been shown to perform well as an estimator.

Finally for completeness, we gather from \cite{jiang2016variational} the various integral formulas required to compute the expectations in closed form with modifications for our multimodal setting. 
\begin{lemma}
Given Gaussian distributions $f(Z) = \mathcal{N}(Z; \mu_1, \sigma_1 \mathbf{I})$ and $g(Z) = \mathcal{N}(Z; \mu_2, \sigma_2 \mathbf{I})$
\begin{equation}
        \int_{\mathbb{R}^l} f(Z) \log g(Z)\, dZ = -\frac12 \sum_j \log 2 \pi \sigma_{2,j}^2 + \frac{\sigma_{2,j}^2}{\sigma_{1,j}^2} + \frac{(\mu_{1,j} - \mu_{2,j})^2}{\sigma_{2,j}^2}.
\end{equation}
\label{intformula1}
\end{lemma}
\begin{lemma}\begin{equation}
    \mathbb{E}_{q(Z,c|X_1,...,X_D)}\left[\log p(Z|c)\right] =  \sum_c q(c|X_1,...,X_D) \int_{\mathbb{R}^l} q(Z|X_1,...,X_D) \log p(Z|c) dZ,
\end{equation}
where the integrand may be computed from Lemma \ref{intformula1}.
\end{lemma}
\begin{lemma}
\begin{equation}
\mathbb{E}_{q(Z,c|X_1,...,X_D)}\left[\log p(Z|c)\right] = \int_{\mathbb{R}^l} q(Z|X_1,...,X_D) dZ \sum_c q(c|X_1,...,X_D) \pi_c,
\end{equation}
\begin{equation*}
= \sum_c q(c|X_1,...,X_D) \pi_c.
\end{equation*}
\end{lemma}
\begin{lemma}
\begin{equation}
    \mathbb{E}_{q(Z,c|X_1,...,X_D)}\left[\log q(Z|X_1,...,X_D)\right] =  \sum_c q(c|X_1,...,X_D) \int_{\mathbb{R}^l} q(Z|X_1,...,X_D) \log q(Z|X_1,...,X_D) dZ,
\end{equation}
\begin{equation*}
    =  \int_{\mathbb{R}^l} q(Z|X_1,...,X_D) \log q(Z|X_1,...,X_D) dZ,
\end{equation*}
where the integrand may be computed from Lemma \ref{intformula1}.
\end{lemma}
\begin{lemma}
\begin{equation}
    \mathbb{E}_{q(Z,c|X_1,...,X_D)}\left[\log q(c|X_1,...,X_D)\right] =  \int_{\mathbb{R}^l} q(Z|X_1,...,X_D) dZ \sum_c q(c|X_1,...,X_D) \log q(c|X_1,...,X_D), 
\end{equation}
\begin{equation*}
    =  \sum_c q(c|X_1,...,X_D) \log q(c|X_1,...,X_D).
\end{equation*}
\end{lemma}
For all terms, $q(c|X_1,...,X_D)$ is calculated via the posterior estimator $\gamma$ given in Equation \eqref{eq:gamma_posterior}. 


\end{document}